\documentclass[runningheads]{llncs}
\usepackage[T1]{fontenc}
\usepackage{graphicx}
\usepackage{booktabs}
\usepackage[misc]{ifsym}

\usepackage[ruled,vlined]{algorithm2e}
\usepackage{amsmath, amssymb}

\usepackage{mwe}
\usepackage{booktabs,multirow,array,graphicx}
\usepackage[table]{xcolor}
\usepackage[normalem]{ulem} 
\usepackage{booktabs,multirow,array,graphicx}
\usepackage[dvipsnames]{xcolor}
\usepackage[normalem]{ulem} 
\usepackage{adjustbox}      
\usepackage{subcaption}
\usepackage{amsmath}

\newcommand{\best}[1]{\textbf{\textcolor{green!45!black}{#1}}}
\newcommand{\second}[1]{\textcolor{red!70!black}{#1}}
\newcommand{\third}[1]{\textcolor{blue!85!black}{#1}}

\usepackage[margin=1.54in]{geometry}
\usepackage[spaces,hyphens]{xurl}
\usepackage{amssymb}
\usepackage{textcomp}

\usepackage{hyperref}

\newcommand{\fref}[1]{Fig.~\ref{#1}}

\usepackage{enumitem} 
\begin{document}

\title{Prototype Fusion: A Training-Free Multi-Layer Approach to OOD Detection}


\author{Shreen Gul\inst{1} \and
Mohamed Elmahallawy\inst{2}
\and  Ardhendu Tripathy\inst{3} \and Sanjay Madria\inst{1} } 


\institute{Missouri University of Science and Technology, Rolla, MO 65401, USA  \email{\{sgchr,madrias\}@mst.edu}
\and
Washington State University, Richland, WA 99354, USA  \\\email{mohamed.elmahallawy}@wsu.edu \thanks{To appear in Proceedings of PAKDD 2026: {\em The} $30^{th}$ {\em Pacific-Asia Conference on Knowledge Discovery and Data Mining.}}
\and
\email{ardhendutr@gmail.com}}

\maketitle              

\begin{abstract}

Deep learning models are increasingly deployed in safety-critical applications, where reliable out-of-distribution (OOD) detection is essential to ensure robustness. Existing methods predominantly rely on the penultimate-layer activations of neural networks, assuming they encapsulate the most informative in-distribution (ID) representations. In this work, we revisit this assumption to show that intermediate layers encode equally rich and discriminative information for OOD detection. Based on this observation, we propose a simple yet effective model-agnostic approach that leverages internal representations across multiple layers. Our scheme aggregates features from successive convolutional blocks, computes class-wise mean embeddings, and applies $L_2$ normalization to form compact ID prototypes capturing class semantics. During inference, cosine similarity between test features and these prototypes serves as an OOD score—ID samples exhibit strong affinity to at least one prototype, whereas OOD samples remain uniformly distant. Extensive experiments on state-of-the-art OOD benchmarks across diverse architectures demonstrate that our approach delivers robust, architecture-agnostic performance and strong generalization for image classification. Notably, it improves AUROC by up to 4.41\% and reduces FPR by 13.58\%, highlighting multi-layer feature aggregation as a powerful yet underexplored signal for OOD detection, challenging the dominance of penultimate-layer-based methods. Our code is available at: \url{https://github.com/sgchr273/cosine-layers.git}.

\end{abstract}
\section{Introduction}
Neural networks  (NNs) have 
remarkable ability to perform complex tasks such as classification, detection, and segmentation with high accuracy~\cite{gul2024lplgrad}. However, despite their success, they exhibit a critical flaw — a tendency to be overconfident when presented with samples that lie outside the training distribution. This overconfidence can lead to catastrophic failures in safety-critical applications such as autonomous driving and medical diagnostics \cite{ammar2023neco}. Hence, it is imperative to design effective methods for out-of-distribution (OOD) detection to ensure the robust deployment of NNs in such domains~\cite{gul2024fishermask}.

A large body of work has focused on {\em leveraging signals from the penultimate layer of NNs} for OOD detection. For instance, Guan et al.~\cite{guan2024exploiting} compute mean feature vectors in the penultimate feature space and exploit the observation that in-distribution (ID) samples exhibit higher similarity to these means than OOD samples. Similarly, KNN~\cite{sun2022out} measures the distance between test and training features, classifying a sample as OOD if the distance exceeds a threshold. NNGuide~\cite{park2023nearest} refines OOD scores by averaging cosine similarities to the $k$ nearest ID neighbors in feature space, while NCI~\cite{liu2025detecting} defines a proximity score based on the norm of the projection of a centered feature onto its class weight vector, with larger norms indicating ID-like behavior. Although these methods achieve competitive results, they primarily depend on the penultimate representation and often fail to capture the richer class-level structure distributed across intermediate layers. As a result, their performance degrades substantially when faced with far-OOD samples~\cite{dong2021neural}.

Recent work shows that informative OOD signals are {\em not restricted to the penultimate layer} but also appear in {\em intermediate representations}. ESOOD \cite{wang2024efficient}, for instance,  trains multiple {\em one-class} SVM detectors at different depths and uses a layer-selection policy for each test sample while using early stopping to terminate inference, but this can be computationally expensive due to the need to handle many layer-specific detectors (e.g., several SVMs per backbone in a ResNet). Ag-EBO \cite{guglielmo2025leveraging} instead regularizes intermediate layers with an energy-based contrastive loss and aggregates several layers into a single OOD score, yet it does not identify which specific layers should be regularized to maximize performance. NMD \cite{dong2021neural} exploits batch-normalization statistics for OOD detection, a direction further supported by \cite{guan2024exploiting,lambert2023multi,jelenic2023out}; however, these methods do not explicitly encode class-specific information, which limits their discriminative power.


To address these limitations, we propose a method that combines {\em fine-grained class information, captured via class prototypes,} with comprehensive multi-layer feature utilization in NNs. Specifically, we construct $L_2$-normalized class prototypes from intermediate layers, which serve as compact, semantically rich representations of ID data. During inference, we compute the cosine similarity between a test sample and each class prototype. For each layer, we retain the maximum similarity across all classes, and a weighted average across layers produces the final confidence score. We observe that ID samples consistently exhibit higher cosine similarity than OOD samples, highlighting cosine similarity as a robust metric for OOD detection. Conceptually, our approach identifies anomalies by analyzing the angles in feature space: ID samples form smaller angles (higher similarity) with class prototypes, whereas OOD samples form larger angles.

\noindent Our main contributions are summarized as follows:
\begin{itemize}[leftmargin=*]
    \item We propose a novel, simple, and training-free OOD detection framework that jointly leverages the class-aware structure of ID data and the network’s internal layers.

    \item Our method exploits geometric relationships in the feature space by computing cosine similarity between class prototypes and intermediate-layer activations. For each layer, the maximum class-wise similarity is retained, and a weighted aggregation across layers yields the final confidence score—enabling robust separation of ID and OOD samples.

    \item Extensive experiments across multiple ID datasets and diverse network architectures demonstrate that our approach consistently enhances OOD detection performance and generalizes effectively across domains. In particular, it improves AUROC by up to 4.7\% and reduces FPR by 13.58\%.
\end{itemize}

\vspace{-0.5cm}
  
\section{Related Work}
\vspace{-0.4cm}
\subsubsection{Neural-collapse-based methods}: A recent thread of work leverages the geometric phenomena of neural collapse (NC) to improve OOD detection. Training-time strategies enforce \emph{feature-space separation}: for example, Wu et al.~\cite{wu2025ncsep} push OOD features into the subspace orthogonal to the ID weight span via a simple orthogonality loss under outlier exposure. Post-hoc alternatives avoid retraining by extracting NC-consistent signals from a trained model: Liu and Qin~\cite{liu2025detecting} score samples using the proximity of centered features to the predicted class weight together with feature norm, while NECO uses NC/PCA structure to distinguish ID from OOD without fine-tuning \cite{ammar2023neco}. Complementary work by Harun et al.~\cite{harun2025controlnc} studies how \emph{controlling} the degree of neural collapse across layers trades off OOD detection and OOD generalization, proposing entropy regularization on the encoder plus an equiangular tight frame (ETF) projector on the head to balance the two objectives. Our method fits naturally within this line: we aggregate intermediate-layer activations into $L_2$-normalized class prototypes and score samples with cosine similarity, thereby capturing discriminative geometry that appears prior to full collapse while remaining post-hoc and model-agnostic.
\vspace{-0.2in}
\subsubsection{Output- and representation-based OOD detection methods}: Another large class of methods transforms a frozen network's outputs or representations into confidence scores. Output-based scores include the \emph{maximum softmax probability} (MSP), which gauges peakiness, and \emph{MaxLogit}, which uses the largest logit directly to mitigate softmax saturation. \emph{Energy}-based scores replace the max with a log-sum-exp (negative free energy) aggregation to leverage the whole logit spectrum and often sharpen ID/OOD separation. Representation-based techniques instead exploit feature geometry: \emph{ViM} decomposes penultimate features into an ID-aligned principal subspace plus a residual to form a virtual logit margin, while residual- or reconstruction-style methods score samples by reconstruction error or distance to a class subspace. Our method complements these approaches by extending geometric scoring beyond the penultimate layer: we build multi-stage, $L_2$-normalized class prototypes from intermediate features and apply cosine similarity, which captures early discriminative cues and yields a simple, effective post-hoc detector that is broadly applicable across architectures.
\vspace{-0.2in}
\subsubsection{Entropy-based OOD detection methods}: The method \cite{rodriguezopazo2025mysteries} extends Maximum Concept Matching using intermediate-layer features from CLIP-like vision–language models.
It applies an entropy-based rule to select and fuse informative layers for training-free OOD detection.
This works best when the backbone has rich, diverse intermediate representations.
Thus it becomes less effective and stable on models with flatter or redundant layers, such as MAE or Perception Encoder.
In contrast, \cite{yang2025eood} models information flow by estimating conditional entropy between consecutive blocks.
It defines a Conditional Entropy Ratio (CER) from ID and jigsaw-based pseudo-OOD images to locate the most discriminative block.
However, EOOD needs jigsaw pseudo-OOD generation and k-NN–based entropy estimation, adding notable overhead.
LaREx \cite{arnez2024latent} proposes uncertainty-based scores LaRED and LaREM using dropout/DropBlock (zMCD) on latent features.
It fits the entropy density of ID samples with kernel density estimation or a Gaussian model.
Yet LaREx still requires Monte Carlo sampling and architecture-specific tuning of the noise layer and its size.
So no universal, computation-light configuration exists, and nontrivial overhead remains.
\vspace{-0.1in}
\section{Problem Statement and Proposed Methodology}
\vspace{-0.1in}
Recent studies indicate that the hidden layers of NNs encode rich, discriminative information about the training data. Motivated by this, we leverage the activations of hidden convolutional layers to design our OOD detection method. To focus on class-specific information and reduce the influence of noise, we compute a mean feature (prototype) for each ID class at every hidden layer.  At test time, each sample is compared to these per-layer prototypes, and the resulting similarities are aggregated across layers. This yields a simple, training-free OOD detector that fuses cosine similarities between samples and class prototypes computed from multiple intermediate layers of a pretrained network. A small ID calibration set is used to construct the prototypes, and OOD scores are obtained by taking the maximum cosine similarity to the prototypes, aggregated across layers (~\fref{fig:main_fig}). The problem statement and details of our method are presented below.
\vspace{-0.1in}
\subsection{Problem Statement}
We consider a pretrained classifier $f \circ h : \mathcal{X} \rightarrow \mathcal{Y}$, where $\mathcal{X}$ denotes the input space (e.g., images), $\mathcal{Y} = \{1,\dots,K\}$ the set of class labels, $h : \mathcal{X} \to \mathbb{R}^d$ a feature extractor (backbone), and $f : \mathbb{R}^d \to \mathcal{Y}$ the classifier head. The model is trained on in-distribution (ID) data drawn from a distribution $P_{\text{in}}$. 

In addition, we are given a small labeled calibration set $\mathcal{D}_{\text{ID}}^{\text{calib}} \subset \mathcal{X} \times \mathcal{Y}$, sampled from $P_{\text{in}}$, which we use to construct class prototypes at a collection of hidden layers $\mathcal{L}$. At test time, a sample $x \in \mathcal{X}$ may originate either from the in-distribution $P_{\text{in}}$ or from an unknown out-of-distribution $P_{\text{out}}$.

Our goal is to design, using only the frozen network $f \circ h$ and the prototype bank built from $\mathcal{D}_{\text{ID}}^{\text{calib}}$, a scalar score function $S(x) \in \mathbb{R}$ that separates ID and OOD samples. The resulting detector is expressed as a decision function $G : \mathcal{X} \to \{0,1\}$,
\begin{equation}
    G(x) =
    \begin{cases}
        1, & \text{if } S(x) \ge \tau \quad (x \text{ is ID}),\\[2pt]
        0, & \text{if } S(x) < \tau \quad (x \text{ is OOD}),
    \end{cases}
    \label{eq:problem_statement}
\end{equation}
where $\tau \in \mathbb{R}$ is a threshold chosen on a validation set.

\subsection{Multi-layer Feature Representations}\label{sec:multi_layer_feat_rep}
Let $f_\theta$ be a pretrained classifier with parameters $\theta$, trained on an ID dataset with $K$ classes. We select a set of internal layers as:
\begin{equation}
    \mathcal{L} = \{\ell_1, \ell_2, \ldots, \ell_L\},
\end{equation}
which may include intermediate convolutional blocks. For an input image $x$, each layer $\ell \in \mathcal{L}$ produces an activation map as:
\begin{equation}
    h_\ell(x) \in \mathbb{R}^{C_\ell \times H_\ell \times W_\ell},
\end{equation}
where $C_\ell$ is the number of output channels while  $H_\ell$ and $W_\ell$ are the spatial height and width of the layer's feature map.
To obtain a fixed-length representation for layer $\ell$, we apply global average pooling over the spatial dimensions $(H_\ell, W_\ell)$. For channel $c \in \{1, \dots, C_\ell\}$, we define
\begin{equation}
    z_\ell(x)_c
    = \frac{1}{H_\ell W_\ell}
    \sum_{i=1}^{H_\ell} \sum_{j=1}^{W_\ell}
    h_\ell(x)_{c,i,j}.
\end{equation}
This yields a channel-wise pooled descriptor $z_\ell(x) \in \mathbb{R}^{C_\ell}$. We then apply $L_2$-normalization to obtain a unit-norm feature vector, which can be given as:
\begin{equation}
    \tilde{z}_\ell(x)
    = \frac{z_\ell(x)}{\lVert z_\ell(x) \rVert_2}
    \in \mathbb{R}^{C_\ell}.
\end{equation}
After this step, each input $x$ is represented by a collection of unit-normalized vectors,
one per selected convolutional layer $\bigl\{ \tilde{z}_\ell(x) \bigr\}_{\ell \in \mathcal{L}}$. 

\begin{figure}[!t]
    \centering
    \includegraphics[width=0.9 \linewidth]{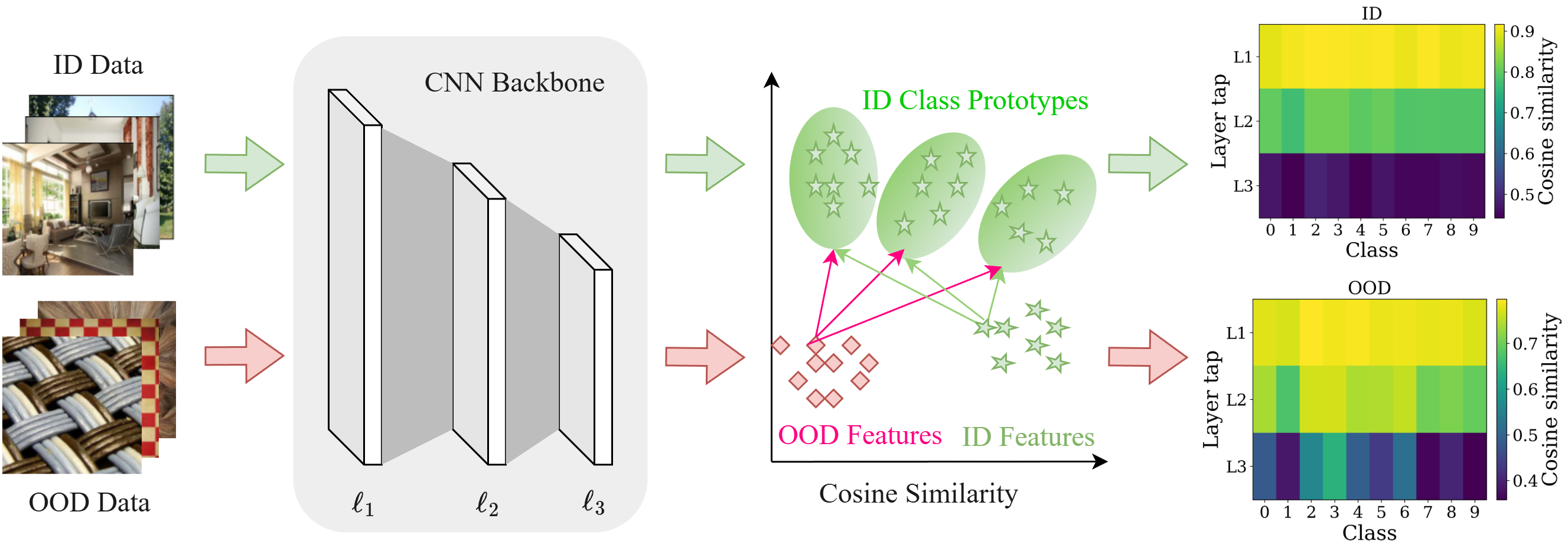}
 \caption{Overview of the proposed training-free OOD detector. A pretrained CNN is tapped at layers \(\ell_1\)–\(\ell_3\). ID calibration images are passed through the network, and per-layer features are global-average-pooled and \(L_2\)-normalized to form \(K\) class prototypes (green regions). For a test image (ID or OOD), we compute cosine similarity between its features and each layer's prototypes (green arrows for ID, red arrows for OOD). At each layer, we retain the maximum class-wise similarity, and these per-layer maxima are averaged to produce an ID affinity score, with \(\text{OODScore} = 1 - \text{affinity}\). The heatmaps on the right show that ID samples yield strong, consistent similarities across layers and classes, whereas OOD samples produce weaker and less coherent similarity patterns.}
\vspace{-0.3cm}
    \label{fig:main_fig}
\end{figure}
\subsection{Class Prototypes in Feature Space}
We construct a prototype representation for each ID class in each selected layer $\ell\in\mathcal{L}$. Let $ \mathcal{D}_{\text{ID}}^{\text{calib}} = \{(x_n, y_n)\}_{n=1}^N,
    \quad y_n \in \{0, \dots, K-1\}$ denotes a labeled calibration set of ID samples. For each layer $\ell \in \mathcal{L}$ and each class $c \in \{0,\dots,K-1\}$, we define the empirical mean feature as:
\begin{equation}
    \bar{p}_{\ell,c}
    = \frac{1}{N_{\ell,c}}
    \sum_{n : y_n = c} \tilde{z}_\ell(x_n),
    \quad
    N_{\ell,c} = \bigl|\{n : y_n = c\}\bigr|.
\end{equation}
We then normalize this class mean to unit length $p_{\ell,c}
    = \frac{\bar{p}_{\ell,c}}{\lVert \bar{p}_{\ell,c} \rVert_2}
    \in \mathbb{R}^{C_\ell}$.  The resulting matrix of class prototypes for each layer $\ell$ can be represented as:
\begin{equation}
    P_\ell =
    \begin{bmatrix}
        p_{\ell,0}^\top \\
        p_{\ell,1}^\top \\
        \vdots \\
        p_{\ell,K-1}^\top
    \end{bmatrix}
    \in \mathbb{R}^{K \times C_\ell}.
\end{equation}
Thus, for every chosen layer $\ell$, we obtain a bank of $K$ unit-norm prototype vectors, one per ID class.
\vspace{-0.2in}
\subsection{Cosine Similarity to Known-Class Manifolds}

Given a test sample $x$, we extract its normalized feature vector $\tilde{z}_\ell(x)$ for each layer $\ell \in \mathcal{L}$, as described in section ~\ref{sec:multi_layer_feat_rep}. We then compute the cosine similarity between $\tilde{z}_\ell(x)$ and each class prototype $p_{\ell,c}$ in that layer:
\begin{equation}
    \text{sim}_\ell(x, c)
    = \tilde{z}_\ell(x)^\top p_{\ell,c},
    \quad c = 0,\dots,K-1.
\end{equation}
\vspace{-0.1in}
Next, for each layer, we retain the maximum similarity over classes as:
\begin{equation}
    m_\ell(x)
    = \max_{c \in \{0,\dots,K-1\}} \text{sim}_\ell(x, c).
\end{equation}
The scalar $m_\ell(x)$ quantifies how well sample $x$ matches any known ID class at layer $\ell$.

\vspace{-0.1in}
\subsection{Cross-layer Aggregation and OOD Scoring}

As different layers capture different forms of structure, we aggregate evidence across all selected layers. We assign a non-negative weight $w_\ell > 0$ to each layer $\ell \in \mathcal{L}$. For a test input $x$, we compute a normalized weighted average of the per-layer match scores as:
\begin{equation}
    s(x)
    = \frac{\sum_{\ell \in \mathcal{L}} w_\ell \, m_\ell(x)}{\sum_{\ell \in \mathcal{L}} w_\ell},
\end{equation}
We interpret $s(x)$ as an ID affinity score, then define the final OOD detection score as:
\begin{equation}
    \text{OODScore}(x)
    = 1 - s(x).
\end{equation}
A higher $\text{OODScore}(x)$ indicates that the sample is less consistent with the set of ID class prototypes across all monitored layers, and is therefore more likely to be OOD.  Algorithm~\ref{alg:oodscore} outlines the scoring procedure.

 \begin{algorithm}[!t]
\caption{Prototype-based multi-layer OOD scoring}
\label{alg:oodscore}
\DontPrintSemicolon
\textbf{Data:} test sample $x$ \;
\textbf{Inputs:} $f_\theta$ (pretrained net), $\mathcal{L}$ (selected layers), 
$\{p_{\ell,c}\}$ (class prototypes), $\{w_\ell\}$ (layer weights) \;
\textbf{Output:} $\text{OODScore}(x)$ \;

\BlankLine

\ForEach{$\ell \in \mathcal{L}$}{
    $h_\ell(x) \leftarrow$ activations of layer $\ell$ on $x$ \;
    $z_\ell(x) \leftarrow \text{GlobalAvgPool}(h_\ell(x))$ \tcp*{$\in \mathbb{R}^{C_\ell}$}
    $\tilde{z}_\ell(x) \leftarrow z_\ell(x) / \| z_\ell(x) \|_2$ \;
    $m_\ell(x) \leftarrow \max_{c} \tilde{z}_\ell(x)^\top p_{\ell,c}$ \tcp*{max cosine sim}
}

$s(x) \leftarrow
\dfrac{\sum_{\ell \in \mathcal{L}} w_\ell \, m_\ell(x)}
      {\sum_{\ell \in \mathcal{L}} w_\ell}$ \;

$\text{OODScore}(x) \leftarrow 1 - s(x)$ \;

\Return $\text{OODScore}(x)$   
\end{algorithm}
\vspace{-0.1in} 

%
%

\vspace{-0.2cm}
\section{Performance Evaluation}
\vspace{-0.2cm}
\subsection{Experimental setup}
\subsubsection{ID \& OOD Datasets.} We evaluate our method on three standard benchmarks for OOD detection: CIFAR-10, CIFAR-100, and ImageNet-1k as the ID datasets, following the experimental protocols of prior work \cite{guan2024exploiting}. For CIFAR-10, we report results under two widely used evaluation protocols: (i) the small OOD benchmark suite consisting of iSUN, LSUN, iNaturalist, Textures, and Places, and (ii) the NECO protocol \cite{ammar2023neco}, where CIFAR-10 (resp. CIFAR-100) serves as the ID dataset and CIFAR-100 (resp. CIFAR-10) together with SVHN are used as OOD datasets. For ImageNet-1k as the ID dataset, we evaluate OOD detection performance against SUN, Textures, Places, and iNaturalist.
\subsubsection{Model Configuration.} Our method involves two key hyperparameters: (i) the weights assigned to the internal layers of the NN, and (ii) the number of samples selected from the ID training data to construct the calibration set ($\alpha$). We conduct an extensive analysis to study the impact of these hyperparameters on the overall OOD detection performance (see results in Figs.~\ref{fig:calib_contribution} and~\ref{fig:layer_weights_regions}).

\vspace{-0.6cm}
\subsubsection{Baselines.} We compare our method against {\em nine} widely used OOD detection baselines. All these methods are \emph{post-hoc}, meaning they operate on pretrained networks without requiring access to model weights, gradients, or retraining losses. This makes them suitable baselines for our method, which is also post-hoc by design. The compared methods include MSP \cite{hendrycks2016baseline}, MaxLogit \cite{hendrycks2019scaling}, Energy \cite{liu2020energy}, Mahalanobis \cite{sehwag2021ssd}, GradNorm \cite{huang2021importance}, NNGuide \cite{park2023nearest}, NECO \cite{ammar2023neco}, ReAct \cite{sun2021react}, and ViM \cite{wang2022vim}.
\vspace{-0.6cm}
\subsubsection{Evaluation Metrics.} We follow the same preprocessing, calibration, and evaluation procedures as in \cite{guan2024exploiting}: a small held-out ID calibration set is used to build prototypes, and all OOD scores are computed post-hoc on pretrained models. Performance is evaluated using the most widely adopted OOD detection metrics: (i) the Area Under the Receiver Operating Characteristic curve (AUROC) and (ii) the False Positive Rate at 95\% True Positive Rate (FPR@95\% TPR). Results are averaged across all OOD datasets unless stated otherwise. Higher AUROC and lower FPR@95\% TPR indicate better OOD detection performance.
\begin{table}[!t]
\centering
\caption{OOD detection results on the {\bf CIFAR-10} dataset. $\uparrow$/$\downarrow$ denote that higher/lower values are better. The \best{first}, \second{second}, and \third{third} best results are highlighted in green, orange, and blue, respectively (values in \%).}
\setlength{\tabcolsep}{1pt}
\resizebox{\linewidth}{!}{ 
\begin{tabular}{ll*{12}{r}} 
\toprule
\multirow{2}{*}{Model} & \multirow{2}{*}{Method} &
\multicolumn{2}{c}{iSUN} & \multicolumn{2}{c}{LSUN} &
\multicolumn{2}{c}{Places} & \multicolumn{2}{c}{iNat} & \multicolumn{2}{c}{Textures} &
\multicolumn{2}{c}{Average} \\
& & AUROC$\uparrow$ & FPR$\downarrow$ &
AUROC$\uparrow$ & FPR$\downarrow$ & AUROC$\uparrow$ & FPR$\downarrow$ &
AUROC$\uparrow$ & FPR$\downarrow$ & AUROC$\uparrow$ & FPR$\downarrow$ &
AUROC$\uparrow$ & FPR$\downarrow$ \\
\midrule
\multirow{12}{*}{\rotatebox{90}{ResNet-18}}
& MSP            & \third{89.79} & \second{47.44} & 94.68 & \third{15.26} & \third{89.66} & \third{48.32} & 89.59 & 44.33 & 87.18 & 63.96 & 90.75 & 39.75 \\
& MaxLogit       & \second{90.08} & 52.84 & 94.63 & 25.82 & 89.65 & 55.06 & 89.37 & 51.19 & 86.35 & 73.86 & 90.74 & 46.60 \\
& Energy         & 88.42 & 63.00 & 92.34 & 51.48 & 87.97 & 62.87 & 87.46 & 59.14 & 84.24 & 79.10 & 89.02 & 57.46 \\
& Mahalanobis    & 86.86 & 51.32 & 80.99 & 53.84 & 84.80 & 49.93 & 80.25 & 58.48 & \second{92.64} & 34.99 & 86.61 & 46.26 \\
& GradNorm       & 83.44 & 70.92 & 91.86 & 37.20 & 79.30 & 74.98 & \second{90.81} & \third{44.29} & 81.20 & 82.75 & 86.56 & 57.97 \\
& NNGuide        & 89.45 & 50.12 & \third{95.18} & \second{13.84} & \second{89.82} & \second{45.34} & \third{90.00} & \second{41.89} & 91.23 & 30.33 & \third{91.71} & 32.93 \\
& NECO           & 84.69 & 54.92 & 85.32 & 53.14 & 81.87 & 61.78 & 83.80 & 56.51 & 91.74 & 42.23 & 87.08 & 50.14 \\
& ReAct          & 85.78 & 75.53 & 91.70 & 50.88 & 84.21 & 79.63 & 84.66 & 73.83 & 80.95 & 92.19 & 86.49 & 69.73 \\
& ViM            & 88.52 & 55.16 & 87.28 & 38.36 & 88.13 & 48.88 & 82.90 & 59.05 & \third{93.88} & \third{30.78} & 89.40 & 42.46 \\
& ESOOD          & 96.70 & \best{8.36}  & \second{96.26} & \best{12.00} & 84.51 & 65.92 & 84.40 & 59.85 & 96.30 & \second{18.37} & 92.66 & \second{29.30} \\
& LaREx          & 89.00 & 39.62 & 86.57 & 48.32 & \best{94.70} & \best{22.68} & 89.94 & 45.76 & \best{98.63} & \best{5.83} & \second{91.77} & \third{32.44} \\
& {\bf Ours}     & \best{93.51} & \second{28.50} & \best{96.34} & 17.97 & 91.61 & 34.60 & \best{92.82} & \best{29.23} & 96.99 & 16.06 & \best{95.00} & \best{22.05} \\
\midrule
\multirow{12}{*}{\rotatebox{90}{DenseNet-100}}
& MSP            & 96.13 & 14.18 & 92.64 & 20.07 & 89.94 & \second{33.02} & 88.10 & 40.26 & 90.86 & 31.39 & 91.86 & 26.11 \\
& MaxLogit       & 98.68 & 6.92 & 95.95 & 16.26 & \second{92.53} & 33.27 & \third{87.74} & 50.95 & 91.29 & 42.66 & 93.73 & 27.37 \\
& Energy         & \second{98.77} & \second{6.65} & 96.08 & 16.10 & \best{92.63} & \third{33.26} & 87.73 & 50.95 & 91.33 & 42.78 & 93.80 & 27.31 \\
& Mahalanobis    & 97.12 & 15.90 & \third{96.52} & \third{15.13} & 70.75 & 77.46 & 75.42 & 71.10 & \third{95.97} & \third{25.11} & 88.83 & 36.69 \\
& GradNorm       & 95.91 & 26.78 & 87.71 & 65.63 & 80.30 & 77.32 & 80.70 & 75.72 & 79.09 & 91.21 & 86.35 & 61.91 \\
& NNGuide        & 95.49 & 18.68 & 93.28 & 19.24 & 88.94 & 42.79 & \second{89.81} & \second{34.08} & 92.92 & 24.66 & 92.37 & 26.34 \\
& NECO           & \third{98.71} & \third{6.88} & 96.00 & 16.01 & \third{92.52} & 33.33 & 87.55 & 50.91 & 91.54 & 40.81 & 93.74 & 27.02 \\
& ReAct          & 98.45 & 9.16 & 95.59 & 18.26 & 92.37 & \best{32.28} & 90.31 & \third{37.95} & 92.17 & 34.57 & \third{94.05} & \third{24.94} \\
& ViM            & \best{99.50} & \best{2.25} & \best{98.62} & \best{6.59} & 90.91 & 38.35 & 85.72 & 54.51 & \second{96.76} & \second{19.01} & \second{95.13} & \second{20.68} \\
& ESOOD          & 97.34 & 10.56 & 83.45 & 53.56 & 87.50 & 68.90 & 85.78 & 65.66 & 34.32 & 96.56 & 81.68 & 58.65 \\
& LaREx          & 96.54 & 20.34 & 91.34 & 35.45 & 90.98 & 36.78 & 88.33 & 39.02 & 96.01 & 15.44 & 92.64 & 29.41 \\
& {\bf Ours}     & 98.24 & 9.81 & \second{97.67} & \second{10.73} & 90.07 & 40.08 &  \best{94.14} & \best{23.59} & \best{97.92} & \best{12.41} & \best{96.13} & \best{17.19} \\
\bottomrule
\end{tabular}}
\label{tab:ood-cifar10-no-svhn}\vspace{-2mm}
\end{table}

\begin{figure}[t]
  \centering

  \begin{subfigure}{0.32\textwidth}
    \includegraphics[width=\linewidth]{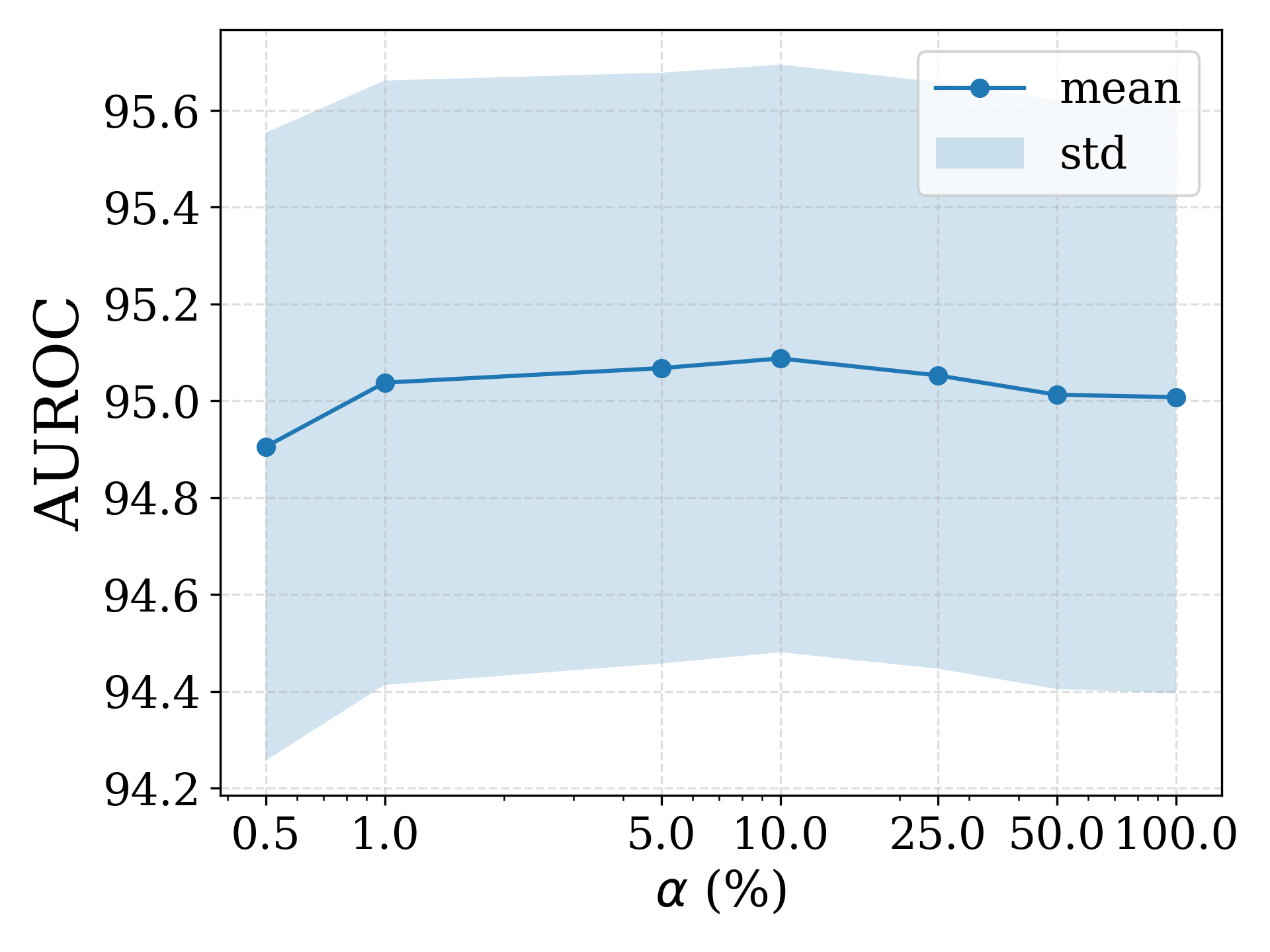}
    \caption{CIFAR-10 AUROC}
    \label{fig:calib_cifar10_auroc}
  \end{subfigure}\hfill
  \begin{subfigure}{0.32\textwidth}
    \includegraphics[width=\linewidth]{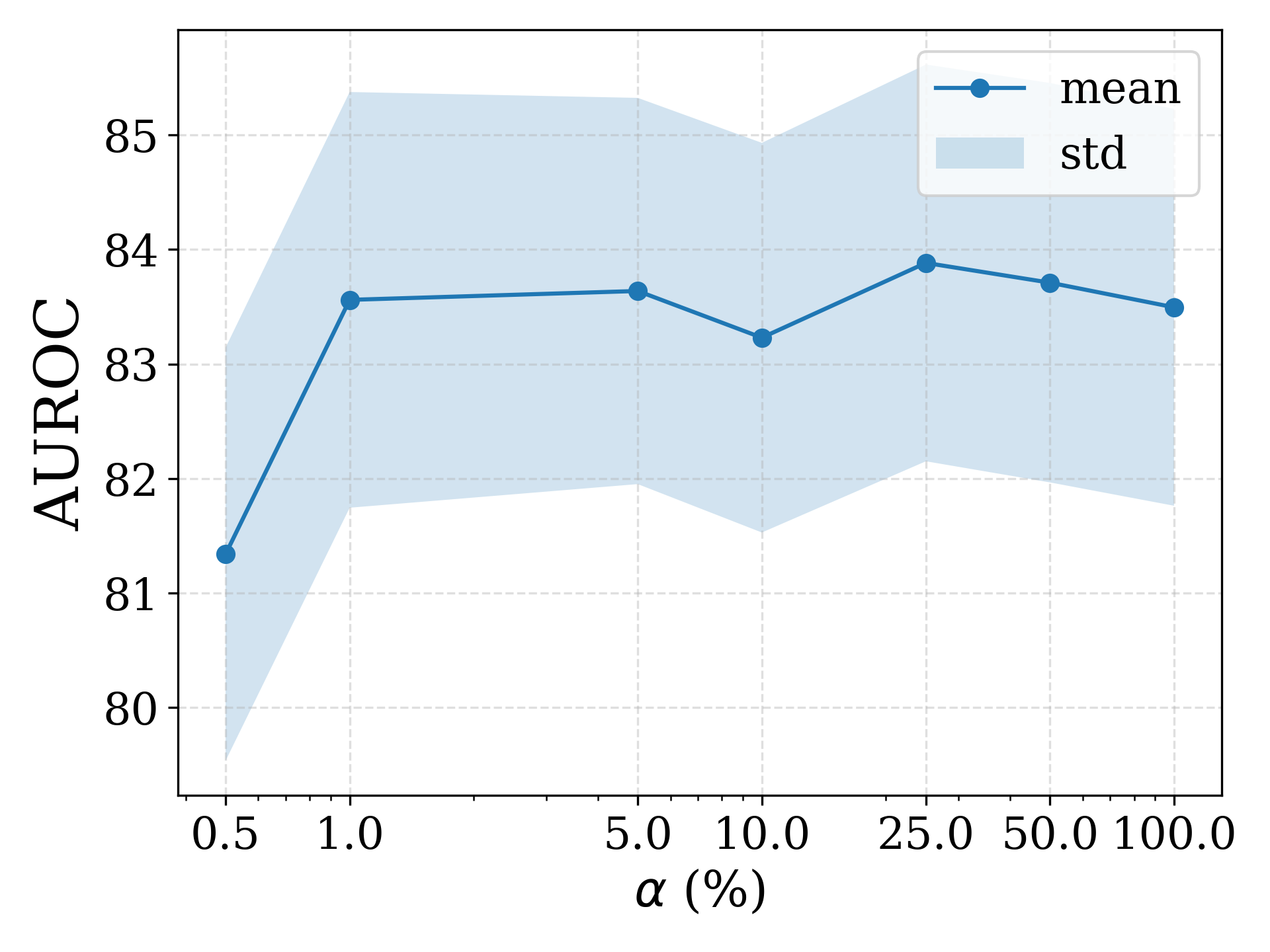}
    \caption{CIFAR-100 AUROC}
    \label{fig:calib_cifar100_auroc}
  \end{subfigure}\hfill
  \begin{subfigure}{0.32\textwidth}
    \includegraphics[width=\linewidth]{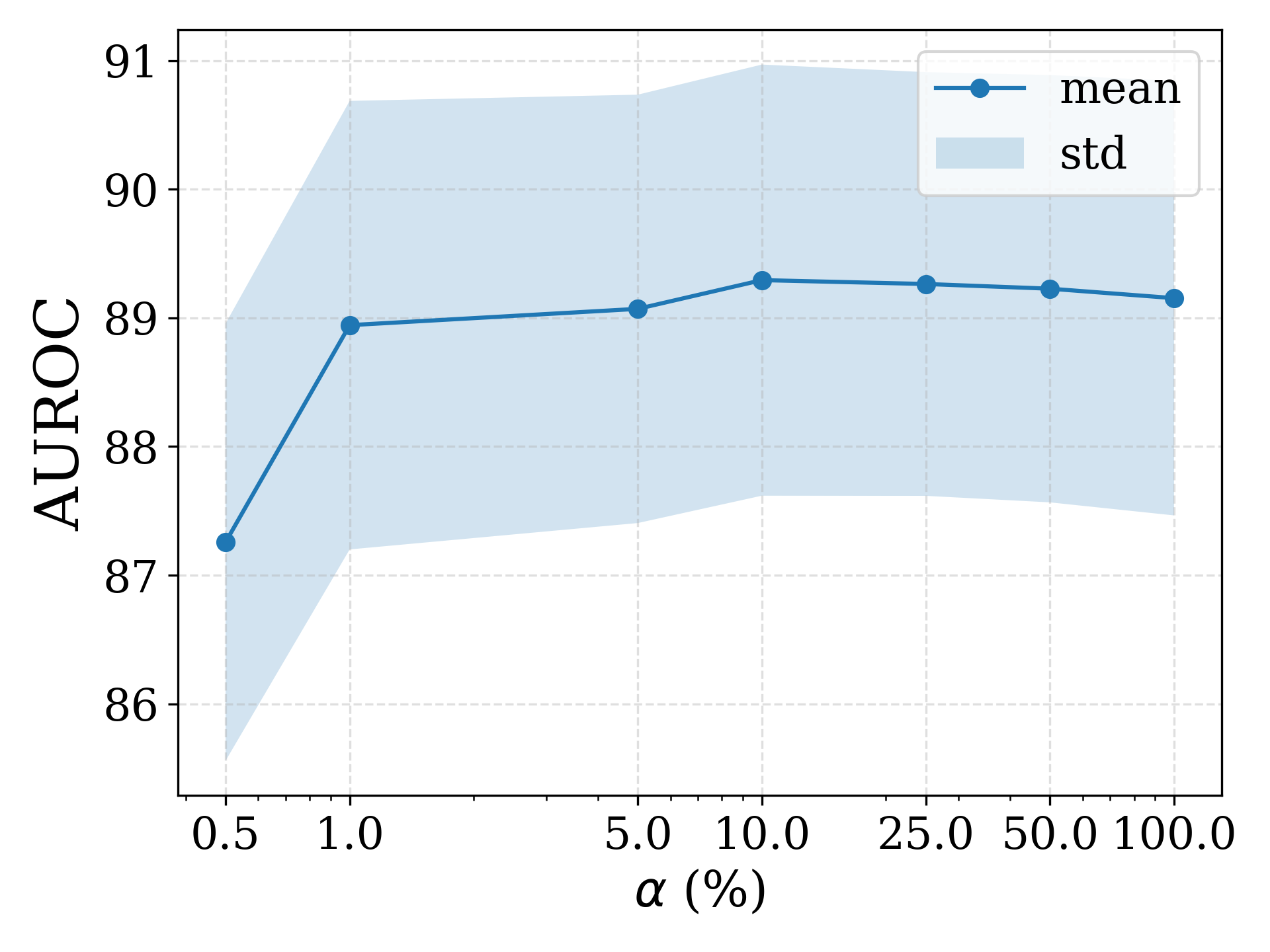}
    \caption{ImageNet-1k AUROC}
    \label{fig:calib_imagenet_auroc}
  \end{subfigure}

  \vspace{0.8em}  

  \begin{subfigure}{0.32\textwidth}
    \includegraphics[width=\linewidth]{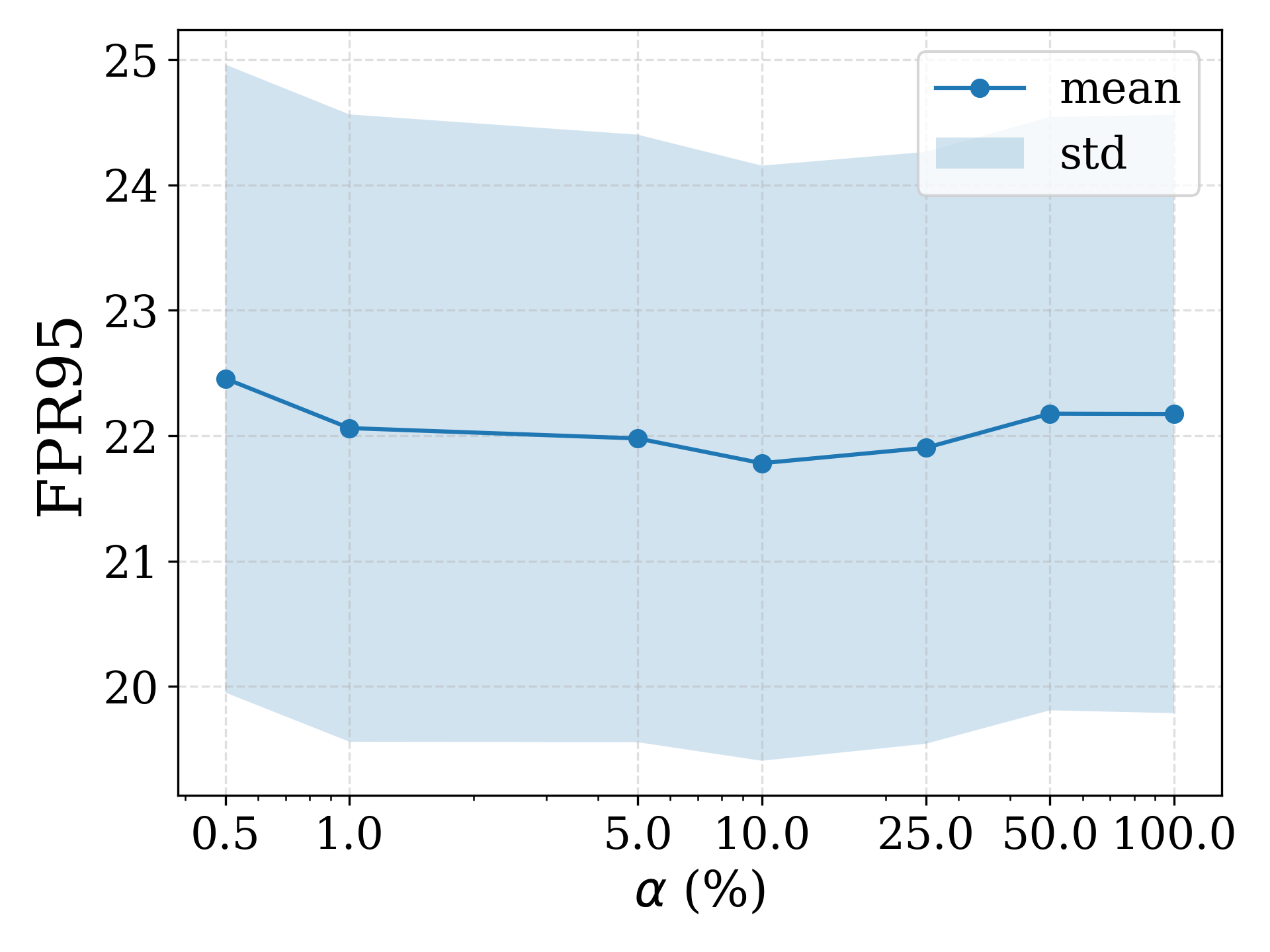}
    \caption{CIFAR-10 FPR}
    \label{fig:calib_cifar10_fpr}
  \end{subfigure}\hfill
  \begin{subfigure}{0.32\textwidth}
    \includegraphics[width=\linewidth]{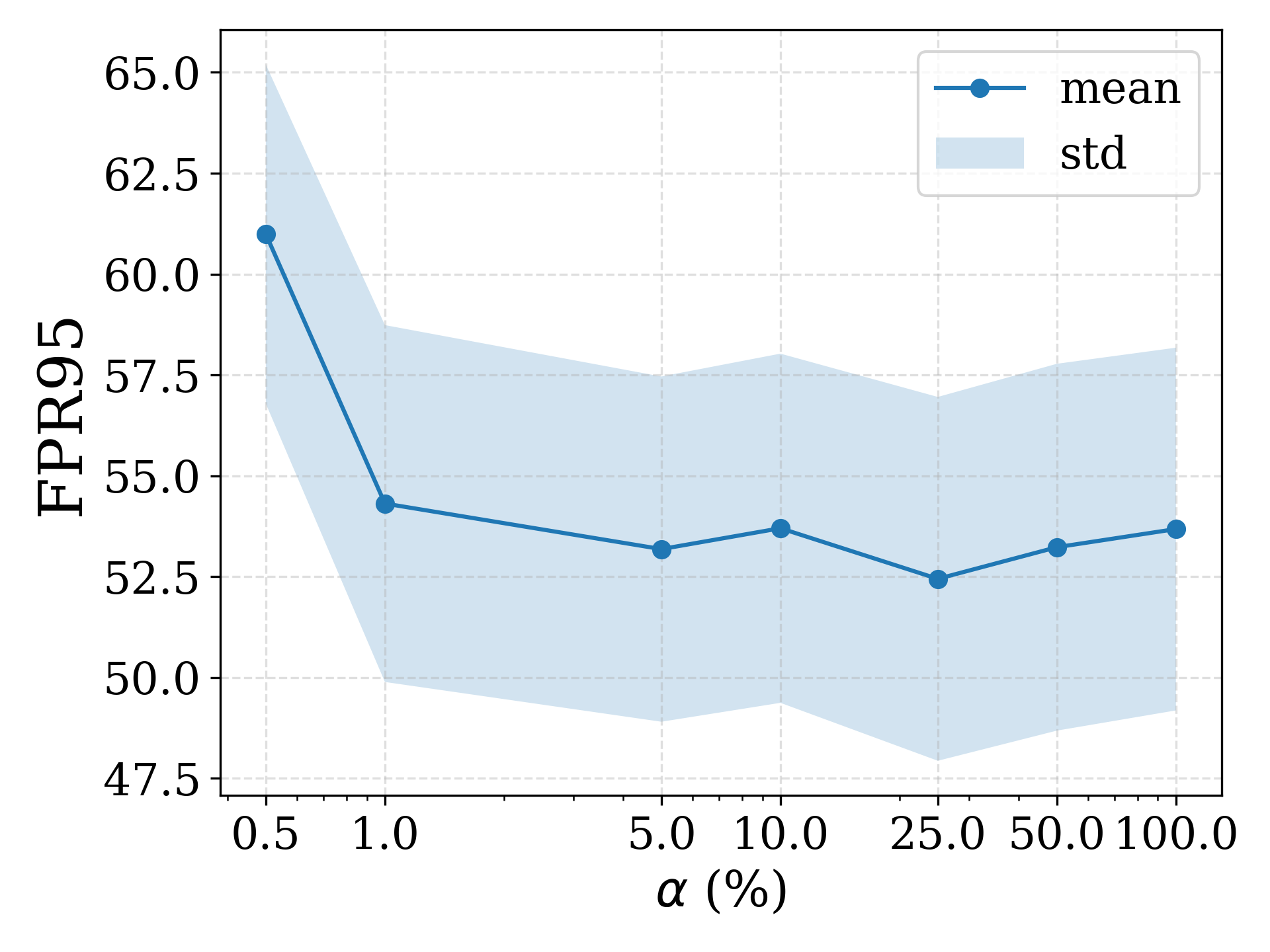}
    \caption{CIFAR-100 FPR}
    \label{fig:calib_cifar100_fpr}
  \end{subfigure}\hfill
  \begin{subfigure}{0.32\textwidth}
    \includegraphics[width=\linewidth]{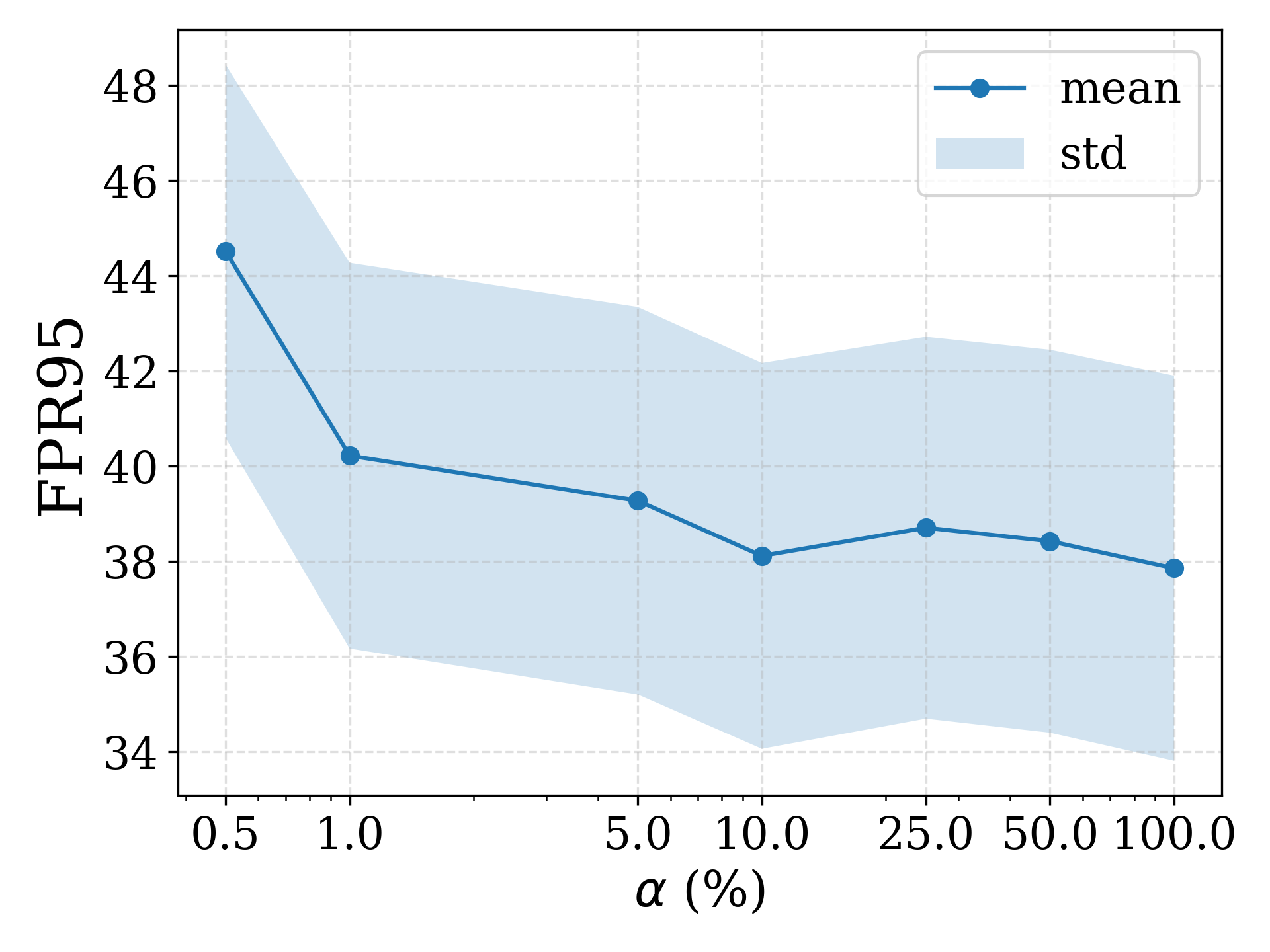}
    \caption{ImageNet-1k FPR}
    \label{fig:calib_imagenet_fpr}
  \end{subfigure}

  \caption{Effect of calibration set size on AUROC (top row) and false positive rate (bottom row) for CIFAR-10, CIFAR-100, and ImageNet-1k.}
  \label{fig:calib_contribution}
\end{figure}



\begin{table}[!t]
\centering 
\small
\caption{OOD detection results where  CIFAR-10 is ID and OOD = CIFAR-100, SVHN) and ID = CIFAR-100 (OOD = CIFAR-10, SVHN). $\uparrow$/$\downarrow$ denote that higher/lower is better. The \best{first}, \second{second}, and \third{third} best results are highlighted in green, orange, and blue respectively (values in \%).}
\setlength{\tabcolsep}{1 pt}
\resizebox{\textwidth}{!}{ 
\begin{tabular}{l l rr rr rr rr rr rr}
\toprule
& & \multicolumn{6}{c}{\textbf{ID = CIFAR-10}} & \multicolumn{6}{c}{\textbf{ID = CIFAR-100}} \\
\cmidrule(lr){3-8} \cmidrule(lr){9-14}
Model & Method &
\multicolumn{2}{c}{CIFAR-100} & \multicolumn{2}{c}{SVHN} & \multicolumn{2}{c}{Average} &
\multicolumn{2}{c}{CIFAR-10} & \multicolumn{2}{c}{SVHN} & \multicolumn{2}{c}{Average} \\
& &
AUROC$\uparrow$ & FPR$\downarrow$ &
AUROC$\uparrow$ & FPR$\downarrow$ &
AUROC$\uparrow$ & FPR$\downarrow$ &
AUROC$\uparrow$ & FPR$\downarrow$ &
AUROC$\uparrow$ & FPR$\downarrow$ &
AUROC$\uparrow$ & FPR$\downarrow$ \\
\midrule
\multirow{12}{*}{\rotatebox{90}{ResNet-18}}
& MSP        & \second{88.76} & \second{46.29} & 93.60 &  \third{19.17} & 91.18 & \third{32.73} & 75.62 & \best{77.26} & 72.92 & 83.01 & 74.27 & 80.14 \\
& MaxLogit   & \third{88.09}  & 53.99         & 94.36 & 20.86          & \third{91.22} & 37.42 & 72.75 & 84.39 & 67.27 & 90.25 & 70.01 & 87.32 \\
& Energy     & 85.04          & 64.90         & 93.69 & 29.20          & 89.37 & 47.05 & 66.38 & 86.81 & 58.26 & 91.68 & 62.32 & 89.25 \\
& Mahalanobis& 82.11          & 59.59         & 94.12 & 29.00          & 88.12 & 44.30 & 70.44 & \second{78.46} & \third{77.92} & \third{61.76} & 74.18 & \third{70.11} \\
& GradNorm   & 76.00          & 77.34         & 92.76 & 37.65          & 84.38 & 57.50 & 27.77 & 97.37 & 22.41 & 98.05 & 25.09 & 97.71 \\
& NNGuide    & \best{89.55}   & \best{41.62}  & 94.59 & 16.09          & \second{92.07} & \second{28.85} & \second{75.82} & \third{79.33} & 76.51 & 68.25 & \third{76.17} & 73.79 \\
& NECO       & 76.86          & 67.72         & \third{95.04} & 32.26   & 85.95 & 49.99 & 39.79 & 95.27 & 45.93 & 91.06 & 42.86 & 93.16 \\
& ReAct      & 80.06          & 82.33         & 91.64 & 46.31          & 85.85 & 64.32 & 28.57 & 95.95 & 36.60 & 92.76 & 32.59 & 94.36 \\
& ViM        & 85.55          & 53.83         & 95.69 & 22.52          & 90.62 & 38.17 & 64.48 & 86.93 & 76.72 & 68.15 & 70.60 & 77.54 \\
& ESOOD      & 84.56          & 62.34         & \second{98.19} & \second{8.67} & \third{91.38} & 35.51 & \best{79.78} & 80.65 & \second{90.19} & \second{35.45} & \second{84.99} & \second{58.05} \\
& LaREx     & 88.20          & 53.45         & 95.32 & 18.50          & 91.76 & 35.98 & 73.11 & 82.45 & 90.45 & 36.09 & 81.78 & 59.27 \\
&  {\bf Ours}        & 87.94          & \third{50.27} & \best{98.94} & \best{5.92} & \best{93.44} & \best{28.10} & \third{74.47} & 79.84 & \best{95.52} & \best{22.31} & \best{85.00} & \best{51.08} \\
\midrule
\multirow{12}{*}{\rotatebox{90}{DenseNet-100}}
& MSP        & 86.78 & \third{47.06} & 93.46 & 17.71 & 90.12 & \third{32.39} & 75.72 & \best{62.15} & 79.82 & 52.74 & 77.77 & 57.45 \\
& MaxLogit   & 86.41 & 59.11         & 96.17 & 14.18 & 91.29 & 36.65 & \second{77.10} & 63.57 & 83.45 & 48.09 & 80.28 & 55.83 \\
& Energy     & 86.44 & 59.12         & 96.28 & 14.12 & 91.36 & 36.62 & \third{76.99} & \third{63.54} & 83.44 & 48.10 & 80.22 & 55.82 \\
& Mahalanobis& 78.99 & 67.97         & 97.18 & 15.42 & 88.09 & 41.70 & 40.51 & 95.53 & 89.23 & 43.25 & 64.87 & 69.39 \\
& GradNorm   & 60.24 & 94.30         & 94.37 & 34.82 & 77.31 & 64.56 & 60.23 & 90.05 & 82.86 & 64.10 & 71.55 & 77.07 \\
& NNGuide    & \third{87.20} & \second{46.60} & 93.78 & 18.59 & 90.49 & 32.60 & 72.79 & 83.51 & 83.04 & 41.86 & 77.92 & 62.69 \\
& NECO       & 86.56 & 58.07         & 96.14 & 14.16 & 91.35 & 36.12 & \best{77.12} & \second{63.50} & 83.85 & 47.19 & 80.49 & 55.35 \\
& ReAct      & \second{87.33} & 51.97 & 95.43 & 17.44 & 91.38 & 34.71 & 75.46 & 68.24 & 85.63 & 41.70 & 80.55 & 54.97 \\
& ViM        & 86.65 & 57.02         & \best{99.29} & \best{3.39} & 92.97 & \second{30.21} & 71.68 & 69.70 & \third{93.66} & \second{28.37} & \third{88.85} & \second{37.74} \\
& ESOOD      & 86.56 & 58.67         & \third{97.89} & \third{11.32} & \third{92.23} & 35.00 & 73.45 & 65.67 & \second{94.56} & \third{30.32} & \best{88.94} & \third{40.20} \\
& LaREx      & 89.34 & 50.15         & 97.99 & 10.15 & \best{93.67} & 30.15 & 75.12 & 70.05 & 93.49 & 29.98 & 84.31 & 50.02 \\
&  {\bf Ours}        & \best{88.34} & \best{45.81} & \second{98.74} & \second{6.53} & \second{93.54} & \best{26.17} & 67.57 & 79.15 & \best{96.33} & \best{17.91} & \second{88.90} & \best{35.11} \\
\bottomrule
\end{tabular}}
\label{tab:cifar10-100-svhn-colored}  \vspace{-2mm}
\end{table}

\begin{table}[!t]
\caption{OOD detection results on the {\bf ImageNet-1k}  dataset. $\uparrow$/$\downarrow$ denote that higher/lower values are better. The \best{first}, \second{second}, and \third{third} best results are highlighted in green, orange, and blue, respectively (values in \%).}
\label{tab:imagenet-ood}
\resizebox{\textwidth}{!}{
\begin{tabular}{ll*{10}{r}}
\toprule
\multirow{2}{*}{Model} & \multirow{2}{*}{Method} &
\multicolumn{2}{c}{SUN} &
\multicolumn{2}{c}{Places} &
\multicolumn{2}{c}{iNat} &
\multicolumn{2}{c}{Textures} &
\multicolumn{2}{c}{Average} \\
& & AUROC$\uparrow$ & FPR$\downarrow$ &
AUROC$\uparrow$ & FPR$\downarrow$ &
AUROC$\uparrow$ & FPR$\downarrow$ &
AUROC$\uparrow$ & FPR$\downarrow$ &
AUROC$\uparrow$ & FPR$\downarrow$ \\
\midrule
\multirow{12}{*}{\rotatebox{90}{ResNet-50}}
& MSP            & 79.25 & 79.48 & \third{77.48} & 78.59 & 84.65 & 62.69 & \third{74.78} & 87.76 & 79.04 & 77.13 \\
& MaxLogit       & 74.78 & 85.81 & 72.64 & 85.90 & 80.25 & 76.47 & 69.25 & 91.74 & 74.23 & 84.98 \\
& Energy         & 50.82 & 89.72 & 50.55 & 90.23 & 51.27 & 87.51 & 48.57 & 94.28 & 50.30 & 90.44 \\
& Mahalanobis    & \third{79.60} & \best{54.97} & 77.26 & \best{65.66} & \third{90.48} & \third{29.89} & \second{87.13} & \second{53.09} & \second{83.62} & \second{50.90} \\
& GradNorm       & 30.01 & 94.09 & 29.56 & 95.31 & 30.72 & 90.89 & 39.25 & 91.06 & 32.38 & 92.84 \\
& NNGuide        & \best{81.14} & \third{73.11} & \best{79.40} & \third{72.72} & \best{97.21} & \best{12.79} & 72.39 & \third{71.51} & \third{82.53} & \third{57.53} \\
& NECO           & 25.13 & 97.40 & 23.98 & 97.67 & 23.85 & 97.13 & 51.02 & 89.28 & 31.00 & 95.37 \\
& ReAct          & 24.10 & 96.45 & 23.08 & 97.37 & 23.87 & 95.26 & 35.14 & 94.52 & 26.55 & 95.90 \\
& ViM            & 51.78 & 90.45 & 51.36 & 90.59 & 82.38 & 55.98 & 49.41 & 93.88 & 58.73 & 82.72 \\
& ESOOD         & 83.45 & 60.67 & 75.67 & 80.38 & 79.76 & 57.87 & 68.87 & 76.89 & 76.94 & 68.95 \\
& LareX          & 82.98 & 61.14 & 85.60 & 70.07 & 91.02 & 30.78 & 96.34 & 20.49 & 88.99 & 45.62 \\
&  {\bf Ours} & \second{80.68} & \second{59.27} & \second{77.68} & \second{66.02} & \second{93.56} & \second{27.25} & \best{96.08} & \best{19.55} & \best{87.00} & \best{43.02} \\
\midrule
\multirow{12}{*}{\rotatebox{90}{RegNet}}
& MSP            & 83.36 & 58.32 & 81.51 & \third{58.44} & 89.84 & 39.13 & 80.74 & 64.79 & 83.86 & 55.17 \\
& MaxLogit       & 79.12 & 63.33 & 75.80 & 61.54 & 85.63 & 44.89 & 73.44 & 69.48 & 78.50 & 59.81 \\
& Energy         & \second{84.05} & \third{58.25} & \third{82.10} & 58.52 & 90.69 & 39.10 & 81.46 & 64.85 & 84.57 & 55.18 \\
& Mahalanobis    & 50.94 & 85.99 & 52.84 & 85.01 & 57.14 & 79.51 & \second{84.98} & 53.62 & 61.48 & 76.03 \\
& GradNorm       & 78.64 & 74.42 & 71.62 & 81.55 & 87.02 & 54.17 & 81.86 & 67.50 & 79.78 & 69.41 \\
& NECO           & 83.17 & 63.49 & 81.02 & 66.29 & 90.30 & 37.00 & 84.19 & 47.98 & 84.67 & \third{53.69} \\
& ReAct          & \best{92.78} & \best{32.86} & \best{90.18} & \best{41.80} & \best{97.02} & \best{14.43} & 91.69 & \second{35.21} & \best{92.92} & \best{31.08} \\
& ViM            & 53.48 & 84.06 & 55.05 & 83.37 & 60.16 & 77.40 & \third{89.62} & 44.16 & 64.58 & 72.25 \\
& NNGuide        & 74.55 & 79.44 & \second{83.72} & \second{55.52} & \third{92.08} & \third{33.31} & 89.60 & \third{35.76} & \third{84.99} & \third{51.01} \\
& ESOOD         & 81.01 & 65.34 & 80.98 & 59.33 & 91.98 & 35.21 & 80.23 & 40.65 & 83.55 & 50.13 \\
& LareX          & 83.03 & 60.92 & 89.75 & 63.72 & 91.43 & 29.14 & 97.76 & 15.61 & 90.49 & 42.35 \\
&  {\bf Ours} & \third{83.83} & \second{56.28} & 80.52 & 63.79 & \second{94.76} & \second{22.45} & \best{98.48} & \best{7.20} & \second{89.40} & \second{37.43} \\
\bottomrule  
 \end{tabular}}\vspace{-2mm}
\end{table}


\vspace{-0.2in}
\subsection{Performance Results}
Tables~\ref{tab:ood-cifar10-no-svhn}, \ref{tab:cifar10-100-svhn-colored}, and \ref{tab:imagenet-ood} report the results of our method compared to all baselines across multiple architectures. In all experiments, we use a uniform weighting of layers. In Table~\ref{tab:ood-cifar10-no-svhn}, for ResNet-18, our method improves AUROC by 2.34\% over ESOOD and 3.66\% over NNGuide, while reducing FPR by 10.88\% and 7.25\%, respectively. For DenseNet-100, our approach increases AUROC by 1.1\% over ViM and 2.5\% over Energy, and reduces FPR by 3.49\% and 4.26\%, respectively. 

In Table~\ref{tab:cifar10-100-svhn-colored} (ResNet-18, ID = CIFAR-10), our method improves AUROC over ESOOD and NNGuide by 0.85\% and 0.69\%, while reducing FPR by approximately 0.75\% and 4.63\%, respectively. For ID = CIFAR-100 on ResNet-18, our method improves AUROC by about 1.99\% over ESOOD and 8.83\% over NNGuide, and reduces FPR by roughly 6.97\% and 19.03\%. On DenseNet-100 with ID = CIFAR-10, our method improves AUROC by 0.5\% over ViM and 1.31\% over MSP, and reduces FPR by 3.5\% and 5.68\% relative to ViM and MSP, respectively. For CIFAR-100, ESOOD edges out our method in AUROC by 0.04\%, but our method achieves a 5.09\% lower FPR. ViM ranks third in AUROC, trailing ours by 0.05\%, and second in FPR, with 2.63\% higher than ours.

In Table~\ref{tab:imagenet-ood} (ResNet-50), our method achieves the best Average score, improving AUROC by 3.38\% over Mahalanobis and 4.47\% over NNGuide, while reducing FPR by 7.9\% and 14.53\%, respectively. For RegNet, our method is the second-best performer after ReAct, with an AUROC lower by 3.52\% and an FPR higher by 6.35\%, whereas ReAct performs poorly on ResNet-50, highlighting that it is extrapolating the signal received by Energy. Compared to the third-place NNGuide, our approach improves AUROC by 4.41\% and reduces FPR by 13.58\%.

\begin{figure}[!t]
    \centering
    \includegraphics[width=0.6 \linewidth]{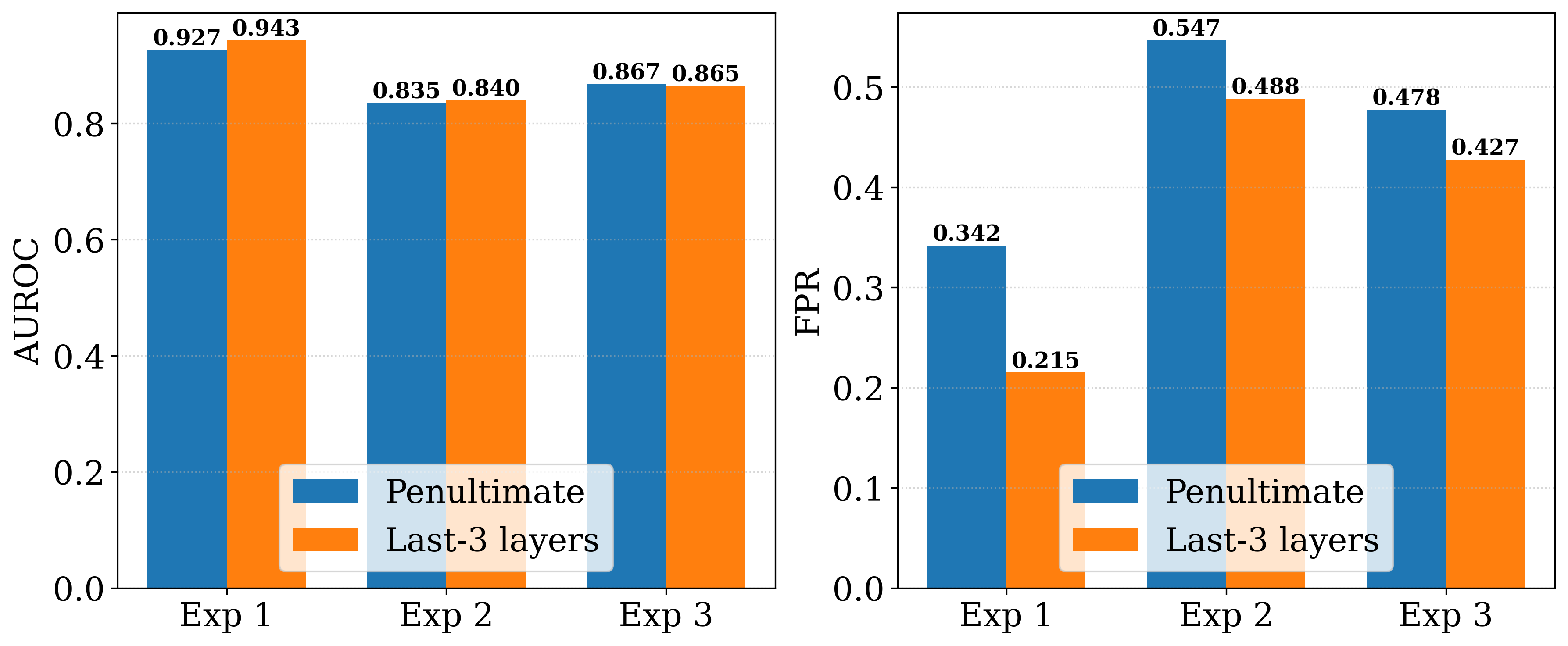}
\caption{AUROC and FPR using the penultimate layer vs. the last three layers across three ID datasets: CIFAR-10 (Exp1), CIFAR-100 (Exp2), and ImageNet-1K (Exp3).}
\vspace{-0.3cm}
    \label{fig:layer_weights_exp}
\end{figure}
\begin{figure}[!t]
    \centering
    \includegraphics[width=0.6 \linewidth]{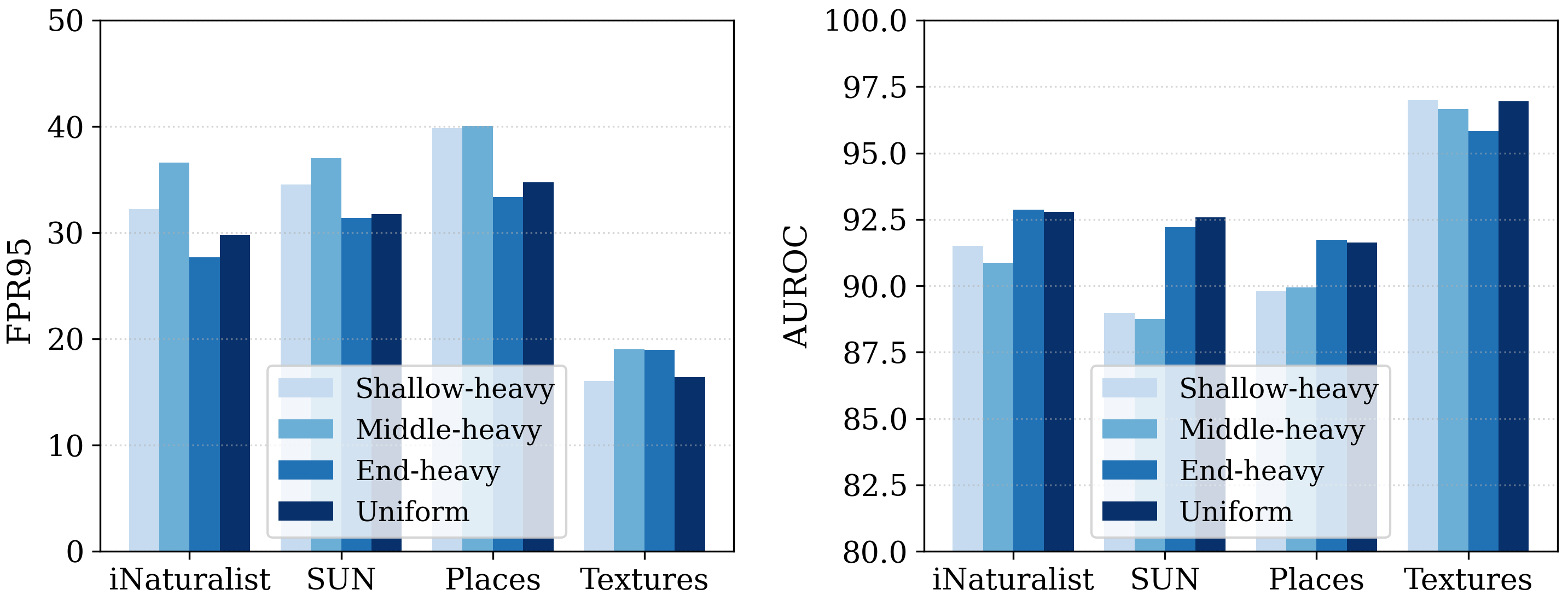}
\caption{Impact of weighting schemes across layers on overall detection performance. }
    \label{fig:layer_weights_regions}
\end{figure}
\vspace{-0.2in}
\begin{figure}[!t]
    \centering
    \includegraphics[width=0.9 \linewidth]{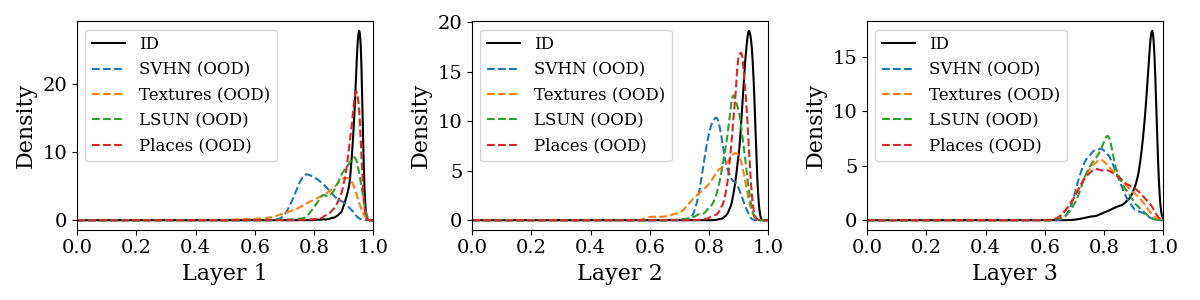}
    \caption{Per-layer cosine similarity scores for the last three layers of ResNet-18 on ID data (CIFAR-10)and four OOD datasets (SVHN, Textures, LSUN, Places).
}
    \vspace{-0.3cm}
    \label{fig:per_layer_cosine}\vspace{-2mm}
\end{figure}
\subsection{Ablation Studies}\label{sec:ablation}
\vspace{-0.1in}
Our method builds an ID representation that captures variations across both classes and hierarchical depths of the backbone network. Accordingly, we analyze two key factors: (i) the stage-level features of the backbone used for prototype construction, and (ii) the size of the calibration set drawn from the ID data to form class prototypes.

\vspace{-0.2in}
\subsubsection{Size of Calibration Set.} We study the impact of calibration set size \((\alpha)\) on performance, averaging results across four OOD datasets (Fig.~\ref{fig:calib_contribution}).  
For \textbf{CIFAR-10}, AUROC improves modestly up to \(\alpha \approx 10\%\) before plateauing, with FPR reaching its minimum near that point.  
For \textbf{CIFAR-100}, most gains occur by \(\alpha \approx 5\%\), with incremental improvement up to \(\alpha \approx 25\%\), where AUROC peaks and FPR is near its minimum. For \textbf{ImageNet-1K}, AUROC rises steadily up to \(\alpha \approx 25\%\) and then levels off, while FPR decreases monotonically.  
Overall, small calibration sets (\(\alpha \approx 5\%\text{--}10\%\)) already perform well, with diminishing yet non-trivial benefits up to \(\alpha \approx 25\%\).
\vspace{-0.2in}
\subsubsection{Cosine Scores across Layers.} Fig.~\ref{fig:per_layer_cosine} visualizes cosine similarity distributions for CIFAR-10 (ID) and four OOD datasets across the last three stages of ResNet-18. The ID scores remain sharply peaked near \(1.0\), while OOD scores cluster lower (\(\sim0.75\!-\!0.90\)) and shift slightly leftward with depth, increasing the separation between ID and OOD samples. This behavior supports cosine similarity as an effective and stable criterion for OOD detection across intermediate layers.

\vspace{-0.2in}
\subsubsection{Weightage of Backbone Stages.}
We evaluate the contribution of different backbone stages under two configurations:
\begin{enumerate}[label={\bf \arabic*.}]
    \item {\bf \em Feature source:} We compare multi-stage features—the final activations from the last three residual stages (before global average pooling)—against using only the penultimate (top) stage. As shown in Fig.~\ref{fig:layer_weights_exp}, aggregating features from the last three stages consistently yields lower FPR and higher AUROC than relying solely on the penultimate stage across all three benchmarks.
    \item {\bf\em Stage weighting:} Within the multi-stage setting, we vary the relative weights assigned to each stage. Fig.~\ref{fig:layer_weights_regions} evaluates four weighting schemes: \emph{shallow-heavy} (largest weight on the earliest stage), \emph{middle-heavy} (largest weight on the middle stage), \emph{top-heavy} (largest weight on the deepest stage), and \emph{uniform} (equal weights). The uniform weighting consistently achieves the strongest performance across datasets.
\end{enumerate}
  
\vspace{-0.3in}

\section{Conclusion and Future Work}
\vspace{-0.1in} 
We proposed a novel post-hoc, training-free OOD detection method that leverages internal network representations and class labels from ID data. By constructing class-specific prototypes from intermediate features and comparing test samples via cosine similarity, our approach captures discriminative cues across multiple layers. Extensive experiments show consistent improvements over existing post-hoc detectors across diverse OOD datasets and ID benchmarks, including large-scale settings such as ImageNet-1K, where it improves AUROC by 4.14\% and reduces FPR by up to 13.85\%. We hope our work underscores the value of internal feature geometry for OOD detection and inspires research that further integrates representation learning with reliable uncertainty estimation.

\vspace{-0.2in}

%
%
%
%
\bibliographystyle{splncs04}
\bibliography{refs}
\end{document}